\documentclass[conference]{IEEEtran}
\IEEEoverridecommandlockouts
\usepackage{float}
\usepackage{cite}
\usepackage{amsmath,amssymb,amsfonts}
\usepackage{algorithmic}
\usepackage{graphicx}
\usepackage{textcomp}
\usepackage{xcolor}

\usepackage{threeparttable}
\usepackage{tabularx}
\usepackage[most]{tcolorbox}
\usepackage[table]{xcolor}
\usepackage{multirow}
\usepackage{hyperref}

\definecolor{prompt_dark_blue}{HTML}{4682b4}
\definecolor{best_value_green}{HTML}{7ee241}

\def\BibTeX{{\rm B\kern-.05em{\sc i\kern-.025em b}\kern-.08em
    T\kern-.1667em\lower.7ex\hbox{E}\kern-.125emX}}
\begin{document}

\title{ M-PACE: Mother Child Framework for Multimodal Compliance\\
}

\author{
\IEEEauthorblockN{Shreyash Verma}
\IEEEauthorblockA{\textit{Intern, Sprinklr} \\
Gurugram, India \\
}
\and
\IEEEauthorblockN{Amit Kesari}
\IEEEauthorblockA{
\textit{Sprinklr AI}\\
Gurugram, India \\
}
\and
\IEEEauthorblockN{Vinayak Trivedi}
\IEEEauthorblockA{
\textit{Sprinklr AI}\\
Gurugram, India \\
}
\and
\IEEEauthorblockN{Anupam Purwar\IEEEauthorrefmark{1}}
\IEEEauthorblockA{
\textit{Sprinklr AI}\\
Gurugram, India \\
}\thanks{\IEEEauthorrefmark{1}Corresponding Author: Anupam Purwar (e-mail: anupam.aiml@gmail.com, https://anupam-purwar.github.io/page/)}
\and
\IEEEauthorblockN{Ratnesh Jamidar}
\IEEEauthorblockA{
\textit{Sprinklr AI}\\
Gurugram, India \\
}

}

\maketitle

\begin{abstract}
Ensuring that multi-modal content adheres to brand, legal, or platform-specific compliance standards is an increasingly complex challenge across domains. Traditional compliance frameworks typically rely on disjointed, multi-stage pipelines that integrate separate modules for image classification, text extraction, audio transcription, hand-crafted checks, and rule-based merges. This architectural fragmentation increases operational overhead, hampers scalability, and hinders the ability to adapt to dynamic guidelines efficiently. With the emergence of Multimodal Large Language Models (MLLMs), there is growing potential to unify these workflows under a single, general-purpose framework capable of jointly processing visual and textual content. In light of this, we propose Multimodal Parameter Agnostic Compliance Engine (M-PACE), a framework designed for assessing attributes across vision-language inputs in a single pass. As a representative use case, we apply M-PACE to advertisement compliance, demonstrating its ability to evaluate over 15 compliance-related attributes. To support structured evaluation, we introduce a human-annotated benchmark enriched with augmented samples that simulate challenging real-world conditions, including visual obstructions and profanity injection.
M-PACE employs a \textit{mother-child} MLLM setup, demonstrating that a stronger parent MLLM evaluating the outputs of smaller \textit{child} models can significantly reduce dependence on human reviewers, thereby automating quality control. Our analysis reveals that inference costs reduce by over 31×, with the most efficient models (Gemini 2.0 Flash as \textit{child} MLLM selected by \textit{mother} MLLM)  operating at \$0.0005 per image, compared to \$0.0159 for Gemini 2.5 Pro with comparable accuracy, highlighting the trade-off between cost and output quality achieved in real time by M-PACE in real life deployment over advertising data.
\end{abstract}

\begin{IEEEkeywords}
Compliance Framework, MLLMs, Structured Model Evaluation, Human-annotated Dataset, Image-Text Analysis, Brand Compliance
\end{IEEEkeywords}

\section{Introduction}

 In the modern digital era, visual media exerts a profound influence on human perception, shaping the reputation and recognition of brands and organizations. The proliferation of multimedia content across online platforms has enabled brands to reach wider audiences, but it also raises the stakes for maintaining consistent compliance with evolving regulatory and ethical standards. Even minor compliance violations in advertisements, such as inappropriate imagery, misleading claims, or unintentional profanities, can have significant repercussions for brand trust and legal standing. As the volume and complexity of media assets grow, manual review processes become increasingly unsustainable, both in terms of cost and scalability. This necessitates the development of automated compliance frameworks capable of evaluating diverse multimodal inputs under real-world conditions.

Traditional compliance pipelines rely on handcrafted rules, object detection models, and OCR modules, which are brittle, domain-specific, and expensive to scale. These systems often require frequent retraining or manual reprogramming to adapt to evolving norms and standards. Moreover, most existing tools focus on narrow objectives, with limited ability to reason semantically or generalize to new tasks. This has led to a fragmented ecosystem of solutions, each tied to a specific modality, objective, or domain.

Multimodal Large Language Models (MLLMs) such as models from GPT \cite{openai2024gpt4o}, Gemini 2.5 \cite{comanici2025gemini25pushingfrontier}, Claude 4 \cite{Claude4:2025}, Llama 4 \cite{llama4blog2025} families offer a compelling alternative by unifying vision and language understanding under a single interface. These models demonstrate strong zero-shot capabilities in layout analysis, image reasoning, visual sentiment inference, and multimodal Q\&A \cite{liu2023visualinstructiontuning}. Importantly, MLLMs offer qualitative advantages over traditional pipelines in terms of maintainability, latency, and cost, enabling rapid compliance \cite{KUMAR2024102783}.

In this paper, we introduce \textbf{M-PACE} - a \textbf{Multimodal Parameter-Agnostic Compliance Engine}, that frames compliance evaluation as a structured reasoning task over multimodal content. While we use advertisement creatives as a representative use case, M-PACE is domain-agnostic and readily generalizes to other domains requiring multimodal content validation.

A novel aspect of \textbf{M-PACE} is the \textit{mother-child} framework, where smaller \textit{child} MLLMs execute compliance tasks such as logo detection or tone consistency, while a larger \textit{mother} MLLM serves as a supervisory agent. The \textit{mother} MLLM periodically evaluates the consistency, accuracy, and robustness of outputs from multiple \textit{child} MLLMs, acting as a judge \cite{li2024llmsasjudgescomprehensivesurveyllmbased}. Such a setup enables dynamic model selection: if a newer or alternative \textit{child} model demonstrates improved performance, the \textit{mother} can recommend or automate a model switch, ensuring continual optimization without retraining or manual auditing. This model-agnostic oversight mechanism makes M-PACE resilient, extensible, and future-ready. 

Following are our key contributions:
\begin{itemize}
    \item We propose \textbf{M-PACE}, a modular and model-agnostic compliance evaluation framework that casts compliance checking as a structured multimodal reasoning task. It supports seamless integration of diverse MLLMs and includes a supervisory mechanism to ensure continual improvement without retraining.
    \item We evaluate the performance of eight SOTA closed-source MLLMs across these attributes, uncovering key strengths and gaps in their zero-shot compliance capabilities, while also benchmarking them on cost and latency, critical for real-world deployment.
    \item We perform a meta-evaluation of the judge (i.e., mother) models using human annotations, employing Cohen’s kappa and accuracy to assess the alignment of MLLMs with expert compliance judgments.
    \item We evaluate MLLM robustness on augmented advertisement images to test operational readiness and uncover potential efficiency gains.
\end{itemize}


\section{Related Work}
Compliance automation has gained significant traction across diverse domains, including finance, food safety, policy interpretation, and brand advertising. This is primarily driven by the maturation of rule-based ML systems, but more recently, by the advent of LLMs. Studies like \cite{hassani2024enhancinglegalcomplianceregulation} demonstrate the potential of LLMs in interpreting structured regulations, while \cite{GHIRANA2012752} highlights the importance of adaptive compliance in enterprise risk management. However, these approaches are largely limited to textual data and fail to generalize to multimodal content. Despite the growing societal and regulatory importance of brand advertisements, compliance evaluation in this space remains vastly underexplored.

The emerging paradigm of LLM-as-a-Judge has gained traction as a scalable alternative to human evaluation across complex tasks. For instance, Zheng et al. \cite{zheng2023judgingllmasajudgemtbenchchatbot} shows that models like GPT-4 and Claude-v1 align closely with expert human evaluators in various NLP tasks. Further studies reinforce the reliability of LLMs in assessment roles \cite{gu2025surveyllmasajudge, chiang2023largelanguagemodelsalternative} \cite{harbola2025knowslmframeworkevaluationsmall} . However, these meta-evaluations are predominantly text-focused, with limited application to visual or multimodal contexts.

Recent work has broadened the scope of multimodal benchmarks. Frank et.al \cite{frank2021visionandlanguagevisionforlanguagecrossmodalinfluence}  explores how modality interactions affect model outputs, a concept relevant to our model-agnostic architecture. MMMU \cite{yue2024mmmumassivemultidisciplinemultimodal} and FaceXBench \cite{narayan2025facexbenchevaluatingmultimodalllms} evaluate MLLMs across multiple disciplines and perception-based tasks. While useful for content understanding, these benchmarks do not address compliance or regulatory evaluation. Similarly, \cite{zhang2025unifiedmultimodalunderstandinggeneration} surveys unified frameworks that combine autoregressive and diffusion-based architectures, paving the way for general-purpose multimodal intelligence. Our work complements these foundational efforts by shifting focus to compliance evaluation, introducing a regulatory lens that current multimodal benchmarks lack.

In terms of datasets, Hussain et al.\cite{hussain2017automaticunderstandingimagevideo} introduced the Video-Ads dataset for analyzing advertisement media across topics, sentiments, and intents. Tencent-AVS \cite{jiang2022tencentavsholisticads} focuses on multimodal scene categorization. Although these datasets offer valuable insight into ad understanding, they largely emphasize classification or generation tasks, not compliance or rule-based evaluation.

Works in persuasion detection and multimodal understanding further contextualize M-PACE. M2P2 \cite{bai2021m2p2multimodalpersuasionprediction} proposes an adaptive fusion model for detecting persuasive intent in videos. More recently,\cite{singla2023persuasionstrategiesadvertisements} categorizes persuasion strategies in advertising, revealing the consistent use of rhetorical and emotional cues. \cite{rizvi2020mdaeads} employs a Multimodal Discourse Analysis (MDA) framework to study how visual elements, such as layout, font, and emphasis convey representational, interactional, and compositional meanings in a small set of e-advertisements. A notable gap still persists in the availability of a scalable, comprehensive, and production-ready solution that is capable of performing end-to-end compliance assessment, supporting both open and closed-source MLLMs, and generalizing across vision and text modalities.


\section{Methodology}
Our evaluation is conducted on the task of brand advertisement compliance, chosen as a representative benchmark for multimodal content validation. This setting involves assessing visual creatives against a diverse set of compliance checks, mentioned in Table \ref{tab:compliance_criteria}. Unlike existing systems that operate purely on explicit visual or textual features, our approach employs an interpretive layer to evaluate whether advertisements conform to both brand objectives and prevailing consumer norms.

\begin{table}[ht]
\caption{Compliance parameters, descriptions and output formats}
\centering
\renewcommand{\arraystretch}{1.1}
\scriptsize 
\begin{tabularx}{\linewidth}{|l|X|l|}
\hline
\textbf{Parameter} & \textbf{Description} & \textbf{Output} \\
\hline
Primary Color      & Prominent color(s) in media & Top-3 colors \\
Logo Detection     & Brand logo in creative           & True/False \\
Logo Position      & Logo location in asset           & Position sector$^{\mathrm{1}}$ \\
Human Presence     & Human figures present            & True/False \\
Face Detection     & Faces present                    & True/False \\
OCR Text           & Recognized text via OCR          & OCR text \\
OCR Overlay Text   & Text overlays from OCR           & OCR segment \\
Headline Text      & Main headline via OCR            & Headline \\
CTA Presence       & Call-to-action phrase            & Phrase(s) detected \\
CTA Position       & Call-to-action phrase location   & Position sector$^{\mathrm{1}}$ \\
Language Detected  & Language of OCR text             & Language(s) \\
Urgent Claim       & Urgent phrases in creative       & Phrase(s) detected \\
Profanity Detection& Profane words in text            & Phrase(s) detected \\
Brand Tone Consistency & Caption tone alignment       & Suggestion \\
Brand Value Consistency & Caption vs. brand values    & Suggestion \\
Brand Positive Phrases  & Positive phrases in text    & Suggestion \\
Brand Negative Phrases  & Negative phrases detected   & Suggestion \\
Grammar Check      & Grammar correctness              & True/False \\
Compliance Check   & Meets content guidelines         & True/False \\
Compliance Consistency & Modifications for compliance & Suggestion \\
Emoji Detection & Presence of emoji & True/False \\
\hline
\end{tabularx}
\label{tab:compliance_criteria}
\\
\raggedright 
{\tiny \hspace{1px}$^{\mathrm{1}}$ Position sector: Center, Top-Left, Top-Right, Bottom-Left, Bottom-Right}
\end{table}

In our setup, each creative is passed through a set of eight \textit{child} MLLMs listed in the Table \ref{tab:openai-google-list-table}, which produce zero-shot predictions for each compliance task in a single forward pass. Zero-shot inference is particularly valuable in real world industry scenarios where latency and cost constraints make multi turn prompting impractical.

\begin{table}[htbp]
\caption{Major OpenAI and Google AI Models}
\centering
\renewcommand{\arraystretch}{1.3}
\scriptsize
\begin{tabular}{|c|l|l|l|l|}
\hline
\textbf{Provider} & \textbf{Model} & \textbf{Input Cost} & \textbf{Output Cost} & \textbf{Temp}\\
\hline
\multirow{4}{*}{OpenAI} 
  & GPT-4.1 & \$2.00 & \$8.00 & 0.0001\\
  & GPT-4.1 Mini & \$0.40 & \$1.60 & 0.0001\\
  & GPT-4o & \$2.50 & \$10.00 & 0.0001\\
  & GPT-4o Mini & \$0.15 & \$0.60 & 0.0001\\
\hline
\multirow{4}{*}{Google} 
  & Gemini 2.5 Pro & \$1.25 & \$10.00 & 0.0001\\
  & Gemini 2.5 Flash & \$0.30 & \$2.50 & 0.0001\\
  & Gemini 2.0 Flash & \$0.15 & \$3.00 & 0.0001\\
  & Gemini 1.5 Pro & \$1.25 & \$5.00 & 0.0001\\
\hline
\end{tabular}
\label{tab:openai-google-list-table}
\vspace{5px}
\\

\raggedright 
{\tiny \hspace{10px} Cost in \$ per 1M tokens}
\end{table}

\begin{figure*}[t]
    \centering
    \includegraphics[width=0.6\textwidth]{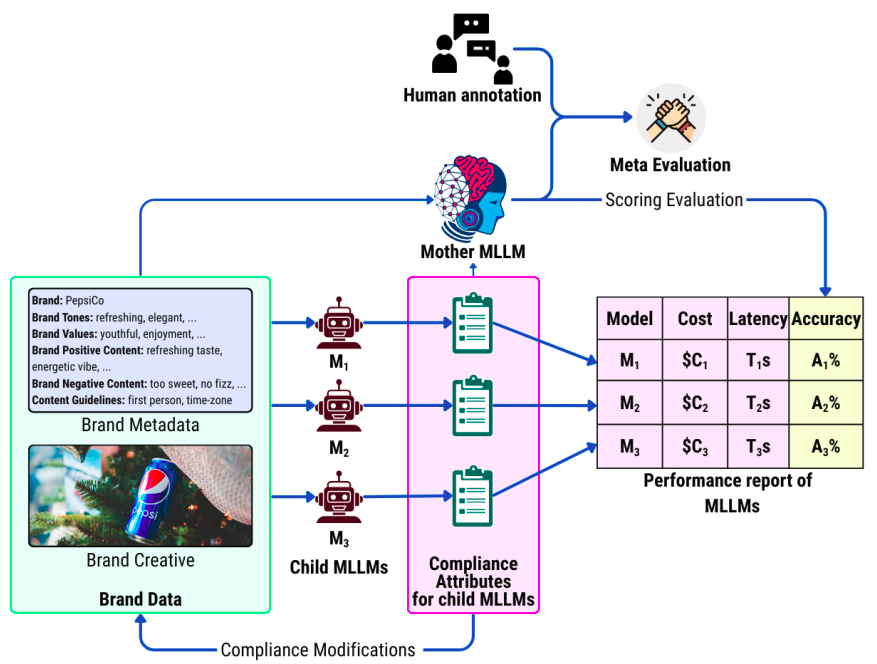}
    \caption{\textbf{Architecture of M-PACE:} \textit{Child} MLLMs independently ingest brand ad creatives to evaluate them against compliance attributes such as brand guidelines, tone, and visual rules, generating compliance attribute reports. These attribute reports are subsequently evaluated for accuracy by a stronger \textit{mother} MLLM using a scoring strategy. Human reviewers provide annotations for benchmarking and participate in the meta-evaluation of the \textit{mother} MLLM. The compliance reports further suggest modifications to ensure alignment with evolving brand or platform policies.}
    \label{fig:architecture}
\end{figure*}

To assess the reliability of these predictions, a supervisory \textit{mother} MLLM (GPT-4.1 and Gemini 2.5 Pro) has been introduced as illustrated in Figure \ref{fig:architecture}. The \textit{mother} MLLM acts as an adjudicator that evaluates and scores the \textit{child} model’s output. 

To evaluate the accuracy of the \textit{mother} model, we utilized ground truth data obtained through human annotation. This setup allows us to compute inter-rater agreement metrics such as Cohen’s kappa and accuracy between the \textit{mother} model’s judgments and human annotations. These metrics help quantify how closely the model aligns with expert assessments across various compliance parameters. Consistently high agreement scores indicate a low error rate, thereby reducing the need for human review and supporting the model’s role as a reliable automated evaluator within an end-to-end compliance system

Before selecting candidate models for our compliance evaluation framework, we reviewed the performance of several open-source MLLMs, including LLaVA 1.5 \cite{liu2023visualinstructiontuning}, BLIP-2 \cite{li2023blip2bootstrappinglanguageimagepretraining}, and more. Although effective on standard benchmarks such as VQAv2 and COCO, they struggle with call-to-actions, spatial logo localization, and recognition of tone, even after prompt or task-specific fine-tuning. Due to these limitations, we exclude open-source MLLMs from further experiments and focus on closed-source models boasting stronger performance in compliance-relevant tasks.

To evaluate the robustness of MLLM based compliance judgments, we further analyze model behavior under a range of controlled visual alterations applied to the input advertisement creatives. These alterations simulate real-world degradations and adversarial scenarios such as changes in brightness, blur, occlusion, rotation, and noise, as well as stylistic disruptions like emoji overlays or the insertion of profane text. The goal is to assess whether MLLM predictions remain consistent when visual inputs are distorted or modified. A complete list of alteration types used in our study is provided in Table \ref{tab:alterations}.

\begin{table}[ht]
\caption{Image alterations applied to test MLLM robustness in compliance evaluation.}
\centering
\renewcommand{\arraystretch}{1.2}
\scriptsize
\begin{tabular}{|l|p{5.4cm}|}
\hline
\textbf{Alteration Type} & \textbf{Description} \\
\hline
Bright & Image brightness increased by 60 intensity units. \\
\hline
Blurred & Blur applied to reduce sharpness. \\
\hline
Dim & Image brightness decreased by 60 intensity units. \\
\hline
Grayscale & image converted to monochrome. \\
\hline
Occluded & logo region partially masked. \\
\hline
Rotated & The image rotated within the range of $[-30^\circ, +30^\circ]$. \\
\hline
Sharp & Edge contrast enhanced. \\
\hline
Gaussian Noise & Random Gaussian noise added. \\
\hline
Salt-and-Pepper Noise & Sparse black and white pixels introduced. \\
\hline
Emoji Overlay & Emojis were overlaid. \\
\hline
Profane Text & Inappropriate or misleading text inserted. \\
\hline
\end{tabular}
\label{tab:alterations}
\end{table}

We initiated prompt development for evaluation with a pilot study on a few human-annotated advertisement creatives. Early iterations focused on aligning model responses with our task-specific output schema, drawing from emerging best practices in prompt engineering such as structural clarity, role specification, and exemplar formatting. Based on systematic failure analysis and qualitative review, we refined the prompt to improve consistency, reduce ambiguity, and ensure faithful content extraction and finalized it as per Figure \ref{fig:brand-compliance-prompt}. We then scaled inference to the full dataset of over 2,700 ads.

In our setup, the primary objective of using the MLLM-as-a-Judge is to ensure high evaluation accuracy, rather than optimizing for cost or latency. Since judgment is performed in a periodic and controlled fashion, we prioritize precision over throughput. To enhance reliability, we decompose complex compliance tasks into separate, focused prompts, allowing the judge model to reason more effectively about specific aspects. Where beneficial, we incorporate few-shot prompting with minimal examples to guide model behavior on nuanced tasks. We also avoid using Likert-style scoring, favoring binary or scalar outputs that offer clearer and more interpretable decisions. This design ensures that the judge model maintains consistency while evaluating \textit{child} MLLM outputs.


\begin{figure}[t]
\centering
\begin{minipage}{0.47\textwidth}
\begin{tcolorbox}[
    colback=white, colframe=black, coltitle=white,
    fonttitle=\bfseries,
    fontupper=\footnotesize,
    title={\large $\circ \circ \circ$},
    enhanced,
    boxrule=1pt,
    colbacktitle=prompt_dark_blue,
    arc=5pt,
    left=2mm,right=2mm,top=2mm,bottom=2mm,
]

\textbf{System Prompt:}

{\scriptsize
\begin{tcolorbox}[colback=gray!5, sharp corners, boxrule=0pt, left=1mm, right=1mm, top=0.5mm, bottom=0.5mm]
Carefully think and reason every response that you give.\\
\textbf{VERY IMPORTANT:} Output result should be in JSON format as told in the user prompt.
\end{tcolorbox}
}

\vspace{1mm}
\textbf{User Prompt:}

{\scriptsize
\begin{tcolorbox}[colback=gray!5, sharp corners, boxrule=0pt, left=1mm, right=1mm, top=0.5mm, bottom=0.5mm]
You are a brand compliance and content safety expert. Given an image, your task is to evaluate it for both brand alignment and public appropriateness. Carefully analyze the content and provide structured feedback that highlights any potential issues requiring attention prior to public release. Your evaluation should be guided strictly by the input details provided below. For each aspect of the assessment, reason through the decision before presenting the final output in the specified structure.

\textbf{Instructions:}

Return the result in the following structure:

\texttt{\{}
\newline \texttt{"Logo Detection": \{ ... \},}
\newline \texttt{"Primary Color": \{ ... \},}
\newline \texttt{"Human Presence": \{ ... \},}
\newline \texttt{"Headline Text": \{ ... \},}
\newline \texttt{"Profanity Detection": \{ ... \},}
\newline \texttt{...}
\newline \texttt{"Compliance Consistency": \{ ... \}}
\newline \texttt{\}}
\end{tcolorbox}
}

\end{tcolorbox}
\end{minipage}
\caption{Brand compliance and content safety prompt used for image analysis.}
\label{fig:brand-compliance-prompt}
\end{figure}

\section{ Dataset}
\begin{figure}[H]
    \centering
    \includegraphics[width=0.35\textwidth]{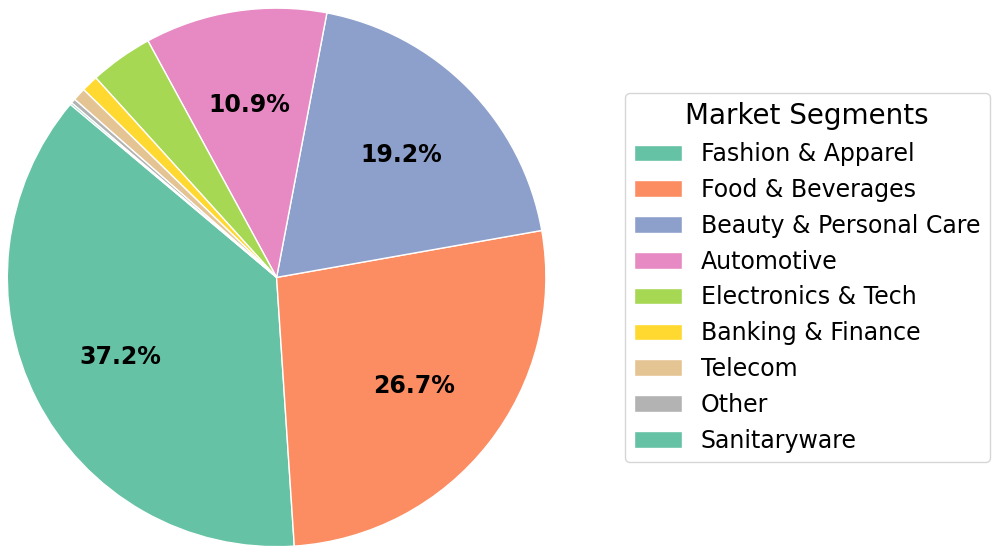}
    \caption{Category-wise distribution across brands.}
    \label{fig:data-dist}
\end{figure}

For our study, we have curated a high-quality subset of over 1,600 image-based advertisements from the Pitts Ad dataset \cite{hussain2017automaticunderstandingimagevideo}. Our selection spans advertisements from over 80 global brands across diverse market segments. To ensure both inter-brand and intra-brand diversity, we have sampled up to 80 distinct creatives per brand, the distribution being illustrated in Figure \ref{fig:data-dist}. Each image is annotated for some compliance dimensions as mentioned in Table \ref{tab:compliance_criteria}. To make sure ads match a brand’s identity, we include brand-specific context in our evaluation. This covers things like the tone a brand usually uses, key values it promotes, and common phrases from its past campaigns. Adding this information helps the system spot when an ad feels off-brand, for example, if the tone doesn’t match or the message feels inconsistent. We use LLMs to automatically gather these brand details from public sources and brand guidelines.

Two trained human annotators have independently labeled images following a standardized brand-specific rubric. A randomly selected subset of 500 images has been double-annotated to assess inter-annotator reliability, resulting in a Cohen’s kappa score of 0.71, indicating substantial agreement. Discrepancies have been resolved through consensus discussions, with escalation to a third adjudicator in rare edge cases. This multi-layered annotation process has ensured high-quality labels suitable for reliable benchmarking and model evaluation.
To evaluate model performance under real-world conditions, we have selected an additional 100 images from the dataset and have applied 11 distinct augmentations to each, as detailed in Table \ref{tab:alterations}, resulting in over 1,100 altered images. This brings our total image Ad set to a total of 2,700 brand creatives.


\section{Experimentation and Evaluation}

\subsection{Evaluation of \textit{child} MLLM}
We conducted evaluations using eight SOTA closed-source Multimodal LLMs mentioned in Table \ref{tab:openai-google-list-table}. Each model was queried using its respective API with a temperature of 0.0001 and a fixed seed value to ensure deterministic outputs. We set the maximum token length to the highest value supported by each multimodal model, along with retries to ensure consistent and complete generation during inference. For the Gemini 2.5 series MLLMs, the reasoning feature remains enabled by default. For each image, we have recorded the results on all of the mentioned compliance parameters in Table \ref{tab:compliance_criteria} as well as calculated the latency, input and output token usages, and total cost spending.

Each compliance parameter is evaluated using task-specific heuristics and brand-aligned expectations. Spatial tasks such as \textit{Logo Position} and \textit{CTA Position} are evaluated against five predefined bins, with the \textit{center} region assigned the highest semantic priority. For parameters such as \textit{CTA Presence} and \textit{Urgent Claim}, we provide a non-exhaustive list of representative phrases within the prompt to guide model understanding. For \textit{primary color detection}, the model is prompted to identify the top three dominant colors present in the image. In tasks such as \textit{tone-based suggestions} and \textit{brand value-based suggestions}, the model generates a text segment intended to align with the brand's core tone and values respectively. \textit{Human Presence} is marked positive if any visible body part indicative of a human is detected, whereas \textit{Face Detection} is restricted to real human faces, excluding caricatures and illustrated avatars. For \textit{OCR Overlay Text}, we apply a threshold where text occupying more than 10\% of the image area is considered significant. \textit{Headline Text} is evaluated by identifying the dominant textual element within the creative. In \textit{Positive Phrase Fix}, we ensure at least one predefined positive phrase is reflected in the generated output, while \textit{Negative Phrase Fix} ensures the excision of all undesired brand-specific terms. Lastly, \textit{Compliance Fix} aggregates multiple linguistic and cultural validation checks, including adherence to time zone conventions, gender inclusivity, audience addressing norms, date formatting, and brand-specific rules.

A key component of our evaluation pipeline is the use of \textbf{MLLM-as-a-Judge (MLLM-J)} to assess the outputs of smaller multimodal models on select compliance parameters that require semantic reasoning or subjective interpretation. For tasks such as \textit{Grammar Check}, \textit{OCR score}, and \textit{Primary Color Detection}, we use GPT-4.1 and Gemini 2.5 Pro as \textit{mother} MLLMs to judge the correctness of candidate outputs. The judge receives the original image, prompt, and model response, and returns a binary or scalar verdict based on contextual alignment. For textual outputs, especially OCR, we employ similarity metrics such as cosine similarity to account for partial correctness, providing a less restrictive alternative to Word Error Rate (WER). This approach builds on recent work showing that LLMs can reliably approximate human evaluation across language and multimodal tasks \cite{ harbola2025knowslmframeworkevaluationsmall}.

\subsection{Meta-evaluation of \textit{mother} MLLM}

To determine the suitability of GPT-4.1 and Gemini 2.5 Pro as \textit{mother} models, we conduct a comparative evaluation against a human-annotated reference set spanning a diverse range of compliance tasks. In our setup, each candidate \textit{mother} model is presented with the original advertisement image, the associated compliance prompt, and the output generated by a smaller \textit{child} MLLM. The \textit{mother} model is then tasked with verifying the correctness of the \textit{child} model’s response.

The judgments produced by each \textit{mother} model are then compared against human annotations using inter-rater agreement metrics such as Cohen’s kappa and accuracy, allowing us to quantify the degree of alignment between MLLM generated evaluations and human expert labels. This comparative evaluation framework is critical for selecting the most reliable and human-aligned \textit{mother} MLLM, which will serve as the backbone for evaluating and scoring outputs from \textit{child} MLLMs in our broader compliance pipeline.

The chosen \textit{mother} model plays a central role in enabling scalable, consistent, and interpretable compliance evaluations without requiring human intervention at each step. By grounding the scoring pipeline in a model that closely mirrors expert-level assessment, we ensure a high degree of trust in evaluations. This setup also supports rapid benchmarking across multiple tasks and models, making it well-suited for evolving real-world compliance needs.

\subsection{Robustness Evaluation}
To evaluate model robustness under realistic input variations commonly encountered in production settings, we have also conducted a series of controlled image augmentations detailed in Table \ref{tab:alterations}. A representative subset of compliance tasks—such as Logo Position, Logo Detection, Human Presence, Face Detection, OCR score, and Primary Color Detection is used to assess output consistency across these altered conditions systematically. In addition, targeted augmentations are applied to evaluate specific capabilities and response accuracy under transformations like image rotation and blur.

For each augmented input, we also log the inference latency and total token usage, enabling a comprehensive analysis of both predictive stability and computational efficiency. This evaluation helps identify models that not only perform well under clean conditions but also maintain reliability and cost-effectiveness when faced with degraded or modified inputs.


\section{Results and Discussion}

\subsection{Evaluation of \textit{child} MLLMs}
While all evaluated models demonstrate competence in basic detection and extraction tasks, significant discrepancies emerge when analyzing more nuanced parameters. Table \ref{tab:accuracy_comparison} presents the exhaustive accuracy results across multiple \textit{child} models and compliance parameters in the evaluation suite. 
GPT-4.1 achieves the highest accuracy in the majority of tasks, specifically, it ranks first in \textbf{15 out of 21} parameters. These include tasks like \textit{Compliance fix}, \textit{grammar check}, \textit{Headline text detection}, to name a few. These findings demonstrate GPT-4.1’s superior alignment with high-level, rule-based content validation and linguistic tasks.

\begin{table*}[ht]
\centering
\caption{Accuracy comparison across \textit{child} models (columns) and compliance parameters (rows), all values in percentage.}
\renewcommand{\arraystretch}{1.2}
\scriptsize
\resizebox{\textwidth}{!}{%
\begin{tabular}{|l|c|c|c|c|c|c|c|c|}
\hline
\textbf{Parameter} & \textbf{GPT-4.1} & \textbf{GPT-4o} & \textbf{GPT-4.1 Mini} & \textbf{GPT-4o Mini} & \textbf{Gemini 2.0 Flash} & \textbf{Gemini 2.5 Flash} & \textbf{Gemini 2.5 Pro} & \textbf{Gemini 1.5 Pro} \\
\hline
Primary Color Detection & 90.85 & 89.30 & 89.05 & 89.37 & 92.23 & 96.18 & \colorbox{best_value_green!20}{\textbf{99.89}} & \underline{98.38} \\

Logo Detection & \colorbox{best_value_green!20}{\textbf{99.57}} & 99.03 & 98.94 & 98.83 & \colorbox{best_value_green!20}{\textbf{99.57}} & 99.56 & 96.34 & 99.26 \\

Logo Position & \colorbox{best_value_green!20}{\textbf{90.28}} & 85.07 & 59.10 & 66.52 & 86.37 & \underline{89.85} & 88.12 & 79.45 \\

Human Presence & \colorbox{best_value_green!20}{\textbf{97.70}} & 97.08 & 97.38 & 95.74 & 97.33 & 95.34 & 94.10 & \colorbox{best_value_green!20}{\textbf{97.70}} \\

Face Detection & 44.57 & 44.88 & 51.37 & 44.41 & 95.53 & \colorbox{best_value_green!20}{\textbf{97.89}} & \underline{97.27} & 94.54 \\

OCR Text Detection & \colorbox{best_value_green!20}{\textbf{98.63}} & \underline{98.30} & 97.59 & 97.67 & 96.78 & 95.13 & 96.80 & 91.91 \\

OCR Overlay Detection & 99.41 & 76.00 & \underline{99.50} & \colorbox{best_value_green!20}{\textbf{99.85}} & 70.98 & 98.02 & 90.19 & 46.11 \\

Headline Text & \colorbox{best_value_green!20}{\textbf{99.58}} & \underline{99.30} & 93.72 & 98.61 & 88.72 & 97.51 & 98.71 & 93.40 \\

CTA Presence & \colorbox{best_value_green!20}{\textbf{99.49}} & \underline{99.30} & 94.76 & 98.31 & 99.00 & 98.88 & 98.80 & 82.90 \\

CTA Position & 86.27 & 80.00 & 59.42 & 49.37 & \underline{88.75} & \colorbox{best_value_green!20}{\textbf{89.64}} & 78.65 & 86.19 \\

Language Detection & \colorbox{best_value_green!20}{\textbf{99.83}} & \underline{99.77} & 99.28 & 98.92 & 99.25 & 99.14 & 99.06 & 92.18 \\

Urgent Claim Detection & \colorbox{best_value_green!20}{\textbf{100.00}} & 99.77 & 99.78 & 99.85 & \colorbox{best_value_green!20}{\textbf{100.00}} & \colorbox{best_value_green!20}{\textbf{100.00}} & \colorbox{best_value_green!20}{\textbf{100.00}} & 98.02 \\

Profanity Detection & \colorbox{best_value_green!20}{\textbf{99.88}} & 99.82 & 99.75 & \colorbox{best_value_green!20}{\textbf{99.88}} & 99.81 & \colorbox{best_value_green!20}{\textbf{99.88}} & 99.44 & 99.13 \\

Tone Consistency & \colorbox{best_value_green!20}{\textbf{98.48}} & 90.52 & 93.11 & \underline{96.60} & 89.82 & 90.46 & 92.94 & 95.14 \\

Value Consistency & \colorbox{best_value_green!20}{\textbf{98.98}} & 95.20 & 97.55 & \underline{97.99} & 95.96 & 96.22 & 97.16 & 95.55 \\

Positive Content Fix & \underline{97.29} & 96.37 & 93.88 & 96.45 & 96.91 & 95.19 & 95.63 & \colorbox{best_value_green!20}{\textbf{98.88}} \\

Negative Content Fix & \colorbox{best_value_green!20}{\textbf{99.92}} & 99.53 & 99.50 & 99.38 & 99.67 & \underline{99.83} & \underline{99.83} & 98.82 \\

Grammar Check & \colorbox{best_value_green!20}{\textbf{99.83}} & \underline{99.77} & 98.73 & 99.54 & 99.24 & 98.02 & 99.31 & 91.01 \\

Compliance Check & \colorbox{best_value_green!20}{\textbf{96.28}} & 87.19 & 90.02 & 87.75 & \underline{91.06} & 70.12 & 70.58 & 90.44 \\

Compliance consistency & \colorbox{best_value_green!20}{\textbf{97.55}} & 96.49 & 96.58 & 97.22 & \underline{97.24} & 93.05 & 95.88 & 75.56 \\

Emoji Detection & 81.81 & 80.80 & \underline{86.17} & 79.10 & 72.54 & 80.39 & \colorbox{best_value_green!20}{\textbf{90.19}} & 73.52 \\

\hline
\textbf{Average} &
94.09 & 91.11 & 90.24 & 90.06 & 93.17 & \colorbox{best_value_green!20}{\textbf{94.29}} & \underline{94.22} & 89.42
\\
\hline
\textbf{Median} &
\colorbox{best_value_green!20}{\textbf{98.63}} & 96.49 & 96.58	& \underline{97.67} & 96.78 & 96.22 & 96.80 & 93.40
\\
\hline

\end{tabular}
}

\label{tab:accuracy_comparison}
\footnotesize
\raggedright 
{\tiny \hspace{1px} Best accuracy percentages per parameter are \colorbox{best_value_green!20}{\textbf{bold and shaded}}, second-best are \underline{underlined}. }
\end{table*}

Despite GPT-4.1’s general superiority, Gemini models exhibit strong performance in specific domains. \textbf{Gemini 2.5 Flash} shows high accuracy in \textit{CTA Position} and performs well across \textit{visual features} like \textit{OCR overlay}. \textbf{Gemini 1.5 Pro} is the only model to match GPT-4.1’s score in \textit{Human Presence}, highlighting its robustness in detecting human related attributes.

Smaller variants, such as GPT-4.1 Mini and GPT-4o Mini, consistently underperform their full-size counterparts across almost all parameters. This aligns with prior observations that models with fewer parameters generally sacrifice depth in contextual reasoning and robustness for cost and speed benefits.

The median accuracy scores reveal interesting patterns in model consistency. GPT-4.1 maintains the highest median performance (98.63\%), suggesting that these models deliver consistently high performance across diverse task types. However, a closer examination reveals a paradoxical pattern: despite GPT models having higher median scores, they show lower average performance compared to some Gemini variants. This discrepancy between median and average indicates that GPT models exhibit extreme performance variability, with exceptional performance in most tasks but critically poor performance in specific domains, notably face detection. In contrast, the Gemini variants show more balanced performance distributions, indicating more consistent performance across all task types.

The results have significant implications for real-world deployment strategies. For applications requiring comprehensive content validation across multiple modalities, GPT-4.1's consistently high performance makes it an optimal choice. However, for vision-heavy applications such as image analysis or visual content moderation, Gemini models offer superior capabilities, particularly in face detection and OCR tasks.
The performance characteristics also suggest that hybrid approaches could be beneficial, leveraging GPT models for text processing and Gemini models for visual tasks. This strategy could optimize both accuracy and computational efficiency while addressing the specific strengths of each model architecture. This reinforces the vision-first architecture of Gemini variants \cite{google2024gemini}, which integrate multimodal vision more deeply into their inference stack.

\subsection{Evaluation of MLLM-as-a-judge - Meta Evaluation}

To understand the reliability of different MLLMs in serving as automated judges for compliance tasks, we assessed their alignment with human annotations across several evaluation categories. This analysis provides insight into each model's strengths and weaknesses as an adjudicator, particularly in tasks that require nuanced interpretation or domain-specific expertise.

\begin{figure}[htbp]  
    \centering
    \includegraphics[width=\linewidth]{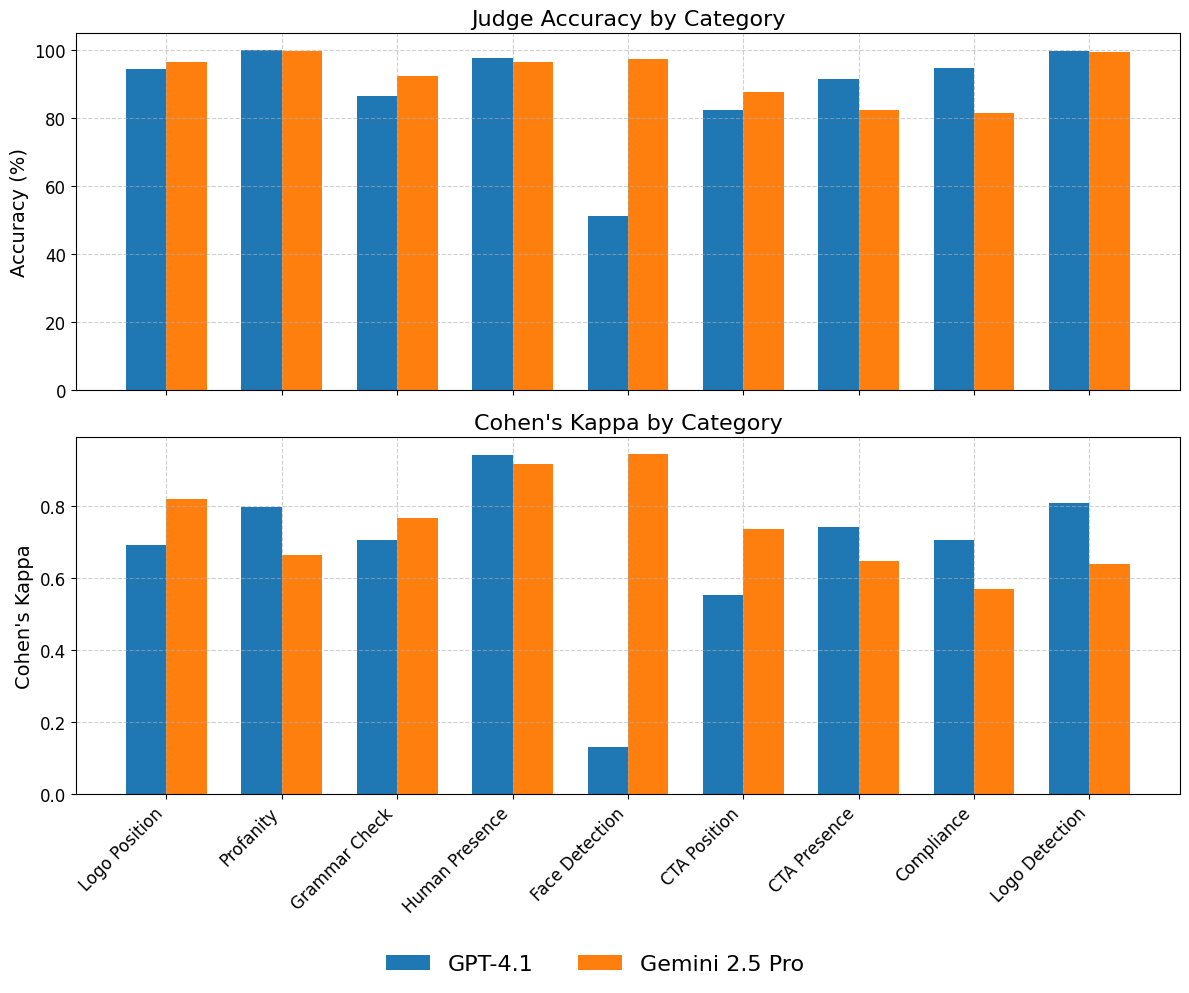} 
    \caption{Comparison of GPT-4.1 and Gemini 2.5 Pro as judges across evaluation categories using Accuracy and Cohen’s kappa.}
    \label{fig:judge_evaluation}
\end{figure}

Figure \ref{fig:judge_evaluation} presents a comparative analysis of the two \textit{mother} MLLMs, GPT-4.1 and Gemini 2.5 Pro, used as automated judges across multiple evaluation categories. GPT-4.1 demonstrates strong performance as a judge, achieving higher Cohen’s $\kappa$ in 5 out of 9 categories. Notably, in tasks like \textit{Face Detection}, GPT-4.1 lags behind significantly in both accuracy (51.28\%) and agreement ($\kappa=0.1303$), indicating difficulty in aligning with human judgments in this domain. In contrast, Gemini 2.5 Pro achieves excellent agreement ($\kappa=0.9460$) and high accuracy (97.40\%), suggesting superior reliability in this specific task.

However, GPT-4.1 shows particularly high agreement and accuracy in categories such as \textit{Human Presence} ($\kappa=0.9430$), \textit{Logo Detection} ($\kappa=0.8086$), and \textit{Profanity Detection} ($\kappa=0.7994$, 99.88\%), outperforming Gemini in each case. Gemini 2.5 Pro shows a slight edge in accuracy over GPT-4.1 in categories like \textit{Grammar Check} and \textit{CTA Position}, outperforming it by approximately 6\% and 5\%, respectively.

\subsection{Performance Evaluation in Augmented Images}

\begin{table}[ht]
\centering
\scriptsize
\caption{Performance metrics under various augmentation variants}
\label{tab:augmented_image_performance}
\begin{tabular}{|l|c|c|c|c|c|c|}
\hline
\textbf{Variant} & \shortstack{\textbf{Logo} \\ \textbf{Pos.}} & \shortstack{\textbf{Logo} \\ \textbf{Det.}} & \shortstack{\textbf{Human} \\ \textbf{Pres.}} & \shortstack{\textbf{Face} \\ \textbf{Det.}} & \textbf{OCR} & \shortstack{\textbf{Color} \\ \textbf{Score}} \\
\hline
Actual         & \colorbox{best_value_green!20}{\textbf{91.0}} & \colorbox{best_value_green!20}{\textbf{99.5}} & \colorbox{best_value_green!20}{\textbf{97.4}} & \colorbox{best_value_green!20}{\textbf{96.9}} & 98.5 & \colorbox{best_value_green!20}{\textbf{97.9}} \\
Bright         & 85.8 & 98.2 & 97.0 & \colorbox{best_value_green!20}{\textbf{96.9}} & 98.5 & 96.5 \\
Blurred        & 90.4 & 99.4 & 95.5 & 95.8 & 97.5 & 97.2 \\
Dim           & 89.5 & 99.2 & 94.6 & 94.7 & 97.00 & 96.7 \\
Gaussian Noise   & 84.0 & 97.7 & 93.7 & 93.6 & 98.2 & 97.5 \\
Grayscale      & 87.7 & \colorbox{best_value_green!20}{\textbf{99.5}} & 92.5 & 92.7 & \colorbox{best_value_green!20}{\textbf{99.2}} & - \\
Rotated        & - & 97.3 & 93.1 & 92.4 & 97.9 & 97.1 \\
Sharp          & 84.9 & 98.8 & 95.2 & 95.8 & 98.3 & \colorbox{best_value_green!20}{\textbf{97.9}} \\
Salt \& Pepper  & 88.6 & 98.3 & 96.1 & 96.2 & 98.2 & 95.2 \\
\hline
\end{tabular}
\normalsize
\end{table}

MLLMs demonstrate a high degree of robustness across a range of image augmentation variants, as shown in Table \ref{tab:augmented_image_performance}. Even under visually challenging conditions, models demonstrate resilience to a variety of visual augmentations. For instance, performance under the \textit{Blurred}, \textit{Dimmed}, and other variants is nearly identical to the unaltered baseline (\textit{Actual}), suggesting minimal impact on perceptual or classification capabilities. These results highlight the models’ capacity to preserve key visual understanding even under degraded input conditions. Missing values in Table~\ref{tab:augmented_image_performance}, marked as “–”, correspond to tasks where specific metrics did not apply to the augmentation applied.


Interestingly, in some cases, augmentation not only preserves performance but marginally improves it. For example, under the \textit{Grayscale} variant, the model surpasses baseline scores in tasks such as OCR Accuracy. This suggests that the removal of color information may help the model de-emphasize irrelevant chromatic cues and focus more effectively on structural and textual elements. This phenomenon highlights the potential of augmentation as a regularization strategy, improving robustness by simplifying the visual input space. Similar findings have been reported in document analysis research, where grayscale conversion has been shown to enhance text recognition by reducing visual noise and distractions \cite{BOUILLON201946}

 \subsection{Latency and Cost}

\begin{figure}[H]
    \centering
    \includegraphics[width=0.45\textwidth]{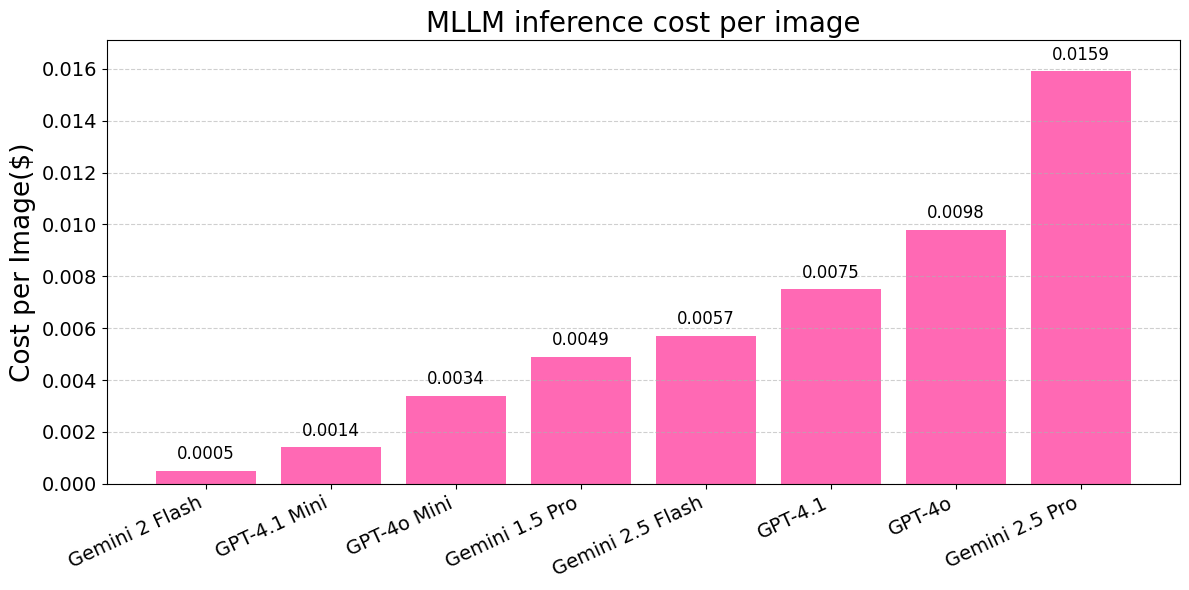}
    \caption{Cost comparison across evaluated models.}
    \label{fig:price-plot}
\end{figure}

\begin{figure}[H]
    \centering
    \includegraphics[width=0.45\textwidth]{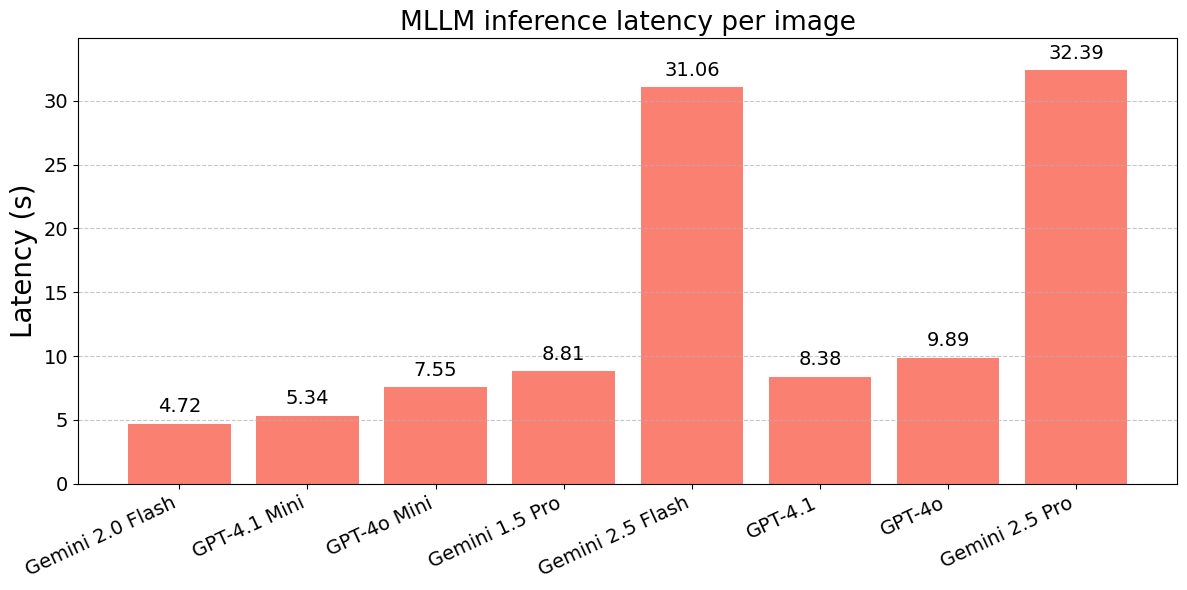}
    \caption{Latency comparison across evaluated models.}
    \label{fig:latency-cost-plot}
\end{figure}

A clear trade-off between cost and latency across the evaluated models exists, as illustrated in Figure \ref{fig:price-plot} and Figure \ref{fig:latency-cost-plot} respectively. Models such as the mini and flash variants offer significantly lower cost per image, making them well-suited for large scale or budget-sensitive deployments. On the other hand, bigger models like {Gemini 2.5 Pro} and {GPT-4o} incur higher processing costs. Notably, the measured latency for Gemini 2.5 models accounts for the default thinking mode.

These trends emphasize the importance of selecting models based on the specified task complexity and operational constraints. While high-capacity models are better suited for tasks requiring semantic reasoning, lightweight models are effective for simpler checks like color detection or presence verification. A hybrid strategy can balance cost, latency, and accuracy.

\section{Conclusion}
This study evaluates multiple SOTA \textit{child} models on a diverse set of visual and compliance-related tasks, benchmarking them along the axes of accuracy, latency, and cost. In addition to direct model evaluation, we also conduct a meta-evaluation of \textit{mother} models to assess how well the judge MLLMs align with human judgments on compliance parameters. Furthermore, we have analyzed model performance under a variety of controlled image augmentations, offering insights into their stability and generalization even in altered conditions. The comparative analysis yields the following key insights:

\begin{itemize}
    \item \textbf{Accuracy and task-specific strengths:} GPT models overall demonstrate superior performance in \textit{text-centric tasks} where linguistic nuance and semantic reasoning are critical, such as grammar checks, profanity detection, and compliance verification. Gemini models show clear superiority in \textit{visual recognition tasks} such as face detection and logo positioning.

    \item \textbf{Inter-Rater reliability and consistency:} While raw accuracy is often close between models, Cohen’s kappa reveals more nuanced differences in consistency. GPT models show higher kappa in text-heavy domains, suggesting more reliable outputs, while Gemini’s higher agreement in visual categories confirms its bias toward spatial features.

    \item \textbf{Latency and cost:} While Gemini 2.5 Pro and Flash demonstrate strong visual accuracy in their thinking mode, they also exhibit the highest inference latency (approximately 32 seconds) and cost per image, which limits their suitability for high-throughput or time-sensitive pipelines. In contrast, lightweight variants such as GPT-4.1 Mini and Gemini 2.0 Flash achieve up to 30$\times$ faster inference with substantially reduced costs, making them more appropriate for use cases where efficiency is prioritized over peak accuracy.

    \item \textbf{Robustness to augmentations:} Models demonstrate robust performance under common augmentations such as blurring, dimming, and slight rotation, with tasks like OCR and human presence detection remaining largely unaffected. 

    \item \textbf{Latency gains:} Certain image transformations, such as grayscale conversion and sharpening, yield measurable reductions in inference time. These augmentations simplify the visual input, enabling faster model processing without compromising performance on key compliance tasks.

    \item \textbf{Failure in specific tasks:} GPT models' major failure in face detection indicates limitations in handling fine-grained visual inputs. Conversely, Gemini models underperform in textual reasoning tasks, indicating a need for improved textual grounding or hybrid fine-tuning.
\end{itemize}

Future work will involve focusing on static video-based advertisement compliance, the core pipeline can be naturally extended to video advertisements. By treating video frames as sequential visual tokens and applying frame sampling techniques. Ultimately, our goal is to evolve the current system into a fully autonomous compliance assistant capable of reasoning, validating, and improving compliance all while minimizing the need for manual intervention and fragmented pipelines.

\section{Acknowledgment}
The authors acknowledge  Amitabh Misra and Yoginkumar Patel for their continuous encouragement to drive innovation through
research in AI. Special thanks to other Sprinklr AI team members for their  support, and constructive feedback throughout the experiment and evaluation  phase of this work.



\bibliographystyle{IEEEtran}
\bibliography{references}



\end{document}